\documentclass[journal,times,onecolumn,12pt]{IEEEtran}
\usepackage{amsfonts,bm,amsmath,graphicx,color,mathtools,balance,algorithmic,algorithm,amssymb,pgfplots}
\usepackage{enumitem}
\usepackage[unicode=true,
 bookmarks=true,bookmarksnumbered=true,bookmarksopen=true,bookmarksopenlevel=1,
 breaklinks=false,pdfborder={0 0 0},backref=false,colorlinks=true,citecolor=blue]
 {hyperref}

\begin{document}
%
\title{Accurate Urban Road Centerline Extraction from VHR Imagery via Multiscale Segmentation and Tensor Voting}
%
%
%

\author{Guangliang Cheng, Feiyun Zhu, Shiming Xiang and Chunhong Pan
\thanks{Guangliang Cheng, Feiyun Zhu, Shiming Xiang and Chunhong Pan are with the National Laboratory of Pattern Recognition, Institute of Automation, Chinese Academy of Sciences (e-mail: \{guangliang.cheng, fyzhu, smxiang and chpan\}@nlpr.ia.ac.cn).}}

%
%

\markboth{}%
{Shell \MakeLowercase{\textit{et al.}}: Bare Demo of IEEEtran.cls for Journals}
%



\maketitle

\begin{abstract}
It is very useful and increasingly popular to extract accurate road centerlines from very-high-resolution (VHR) remote sensing imagery for various applications, such as road map generation and updating etc.  There are three shortcomings of current methods: (a) Due to the noise and occlusions (owing to vehicles and trees), most road extraction methods bring in heterogeneous classification results; (b) Morphological thinning algorithm is widely used to extract road centerlines, while it produces small spurs around the centerlines; (c) Many methods are ineffective to extract centerlines around the road intersections. To address the above three issues, we propose a novel method to extract smooth and complete road centerlines via three techniques: the multiscale joint collaborative representation (MJCR) $\&$ graph cuts (GC), tensor voting (TV) $\&$ non-maximum suppression (NMS) and fitting based connection algorithm. Specifically, a MJCR-GC based road area segmentation method is proposed by incorporating mutiscale features and spatial information. In this way, a homogenous road segmentation result is achieved. Then, to obtain a smooth and correct road centerline network, a TV-NMS based centerline extraction method is introduced. This method not only extracts smooth road centerlines, but also connects the discontinuous road centerlines. Finally, to overcome the ineffectiveness of current methods in the road intersection, a fitting based road centerline connection algorithm is proposed. As a result, we can get a complete road centerline network. Extensive experiments on two datasets demonstrate that our method achieves higher quantitative results, as well as more satisfactory visual performances by comparing with state-of-the-art methods. As another contribution, a new and challenging road centerline extraction dataset for VHR remote sensing images is made and publicly available for further studies.
\end{abstract}

\begin{IEEEkeywords}
 road centerline extraction, multiscale joint collaborative representation (MJCR) $\&$ graph cut (GC), tensor voting (TV) $\&$ non-maximum suppression (NMS), fitting based connection.
\end{IEEEkeywords}

%
\IEEEpeerreviewmaketitle

\section{Introduction}
Accurate road extraction from remote sensing images is an essential preprocessing step for various applications, such as vehicle navigation \cite{journals_tvt_LiCLSN14}, geographic information system (GIS) update \cite{journals_prl_BonnefonDD02} and intelligent transportation system \cite{journals_tits_KumarSR05} etc. However, it is time-consuming, costly and tedious to manually label the road area from the image. Thus, it is desired to find ways to automatically extract road areas from images. Although, recent research on road extraction \cite{journals_prl_Mena03, journals_survey_MFDC03, journals_tgrs_DasMV11, journals_IJR_AE15} have been proposed to address this challenging task, they are far from mature.
\par  According to the resolution of sensors, remote sensing images can be classified as low-, median- and high-resolution images. Roads on the low- and median-resolution images tend to be lines with small width of only one or two pixels. Recent line detection methods, such as dynamic programming and snake algorithms \cite{journals_pers_Gruen97}, template matching \cite{conference_acrs_Park01} and Hough transform \cite{journals_tgrs_DellAcquaG01a}, have focused on the extraction of road networks from these images. Due to recent advances in remote sensors, a large amount of high-resolution images become available, which exhibit more details about the earth surface. It is an urgent task to extract road from high-resolution images for various applications. However, compared with road extraction from low- and median-resolution images, there are a number of difficulties to extract the road area from high-resolution remote sensing imagery. First, small objects can be observed and the images tend to be affected by noise. Thus the spectral signatures of road become more heterogeneous. Second, complex backgrounds and contextual objects, such as trees, buildings and vehicles on the roads, usually appear in the high-resolution images. Finally, there are some road-like objects, such as buildings, rivers and parking lots, that may be misclassified as roads.
\par Most road area extraction methods \cite{journals_prl_ZhangC06, journals_ijrs_Huangxin09} are based on the pixel-based classification method, which brings in heterogeneous results due to the noise and occlusions under vehicles and trees in the VHR remote sensing images. For the road centerline extraction methods, many researchers \cite{journals_ijrs_Huangxin09, journals_jstars_DCNKAS12, journals_lgrs_ShiMWZ14} applied the morphological thinning algorithm, which is both fast and easy to perform. However, this algorithm produces many small spurs, which reduce the correctness and smoothness of the road network. Though some regression based centerline extraction methods \cite{journals_tgrs_ShiMD14, journals_lgrs_MiaoSZW13} are introduced to solve this shortcoming, they are ineffective to extract the centerlines around the road intersections.
\par To overcome the above shortcomings in the existing methods, we propose a novel three-stage based method to extract smooth and complete road centerlines from very-high-resolution (VHR) remote sensing images: 1) Homogeneous road area segmentation; 2) Smooth and correct road centerline extraction; 3) Centerline connection around the road intersections. The proposed method integrates three techniques, that is, multiscale joint collaborative representation (MJCR) $\&$ graph cuts (GC), tensor voting (TV) $\&$ non-maximum suppression (NMS) and fitting based connection algorithm. Specifically, to obtain a homogeneous road area segmentation result, a MJCR-GC based road segmentation algorithm is proposed, which incorporates multiscale contextual features and spatial information. Then, a TV-NMS based centerline extraction algorithm is put forward to gain smooth and correct centerlines. Finally, to well connect the centerlines in the road intersections, a fitting based connection algorithm is introduced.
\par The main contributions of the proposed approach are highlighted as follows:
\begin{itemize}[leftmargin=*]
\item MJCR and GC are combined to obtain a homogenous road segmentation result. In MJCR, a novel road based feature is firstly proposed, which integrates spectral, structural and contextual road characteristics. This feature is in line with the human perception of road recognition.
\item A new TV-NMS based centerline extraction method is introduced to extract the road network. It can not only extract smooth and correct road centerlines, but also connect the nearby discontinuous centerlines due to unconnected regions in the segmentation result.
\item To overcome the ineffectiveness of the existing road centerline methods in the intersection areas, a fitting based connection algorithm is proposed to complete those unconnected centerlines around the road intersections.
\item A new and challenging road centerline extraction dataset is publicly available for further studies. It contains 30 VHR remote sensing images together with the corresponding centerline reference maps.
\end{itemize}
\par The remainder of this paper is arranged as follows. The related road extraction work is systematically reviewed in Section \ref{Section2}. In Section \ref{Section3}, the details of the proposed road area extraction and centerline extraction method are introduced. Experimental evaluations as well as detailed comparisons between our method and state-of-the-art methods are provided in Section \ref{Section4}. Finally, the conclusions will be outlined in Section \ref{Section5}.
\section{Previous Work}
\label{Section2}
\par For VHR images, according to the extracted road results, the existing road extraction approaches can be classified into two classes: 1) {\it road area extraction} methods, 2) {\it road centerline extraction} methods.
\par Road area extraction methods mainly depend on image classification and segmentation. Zhang et al. \cite{journals_prl_ZhangC06} proposed an integrated approach that combines {\it k}-means, fuzzy logical classifier and shape descriptors of angular texture signature. It can separate the roads from the parking lots that have been misclassified as roads. A new method for extracting roads based on advanced directional morphological operators was proposed in \cite{journals_prl_ValeroCBTW10}, in which Path Openings and Path Closings were introduced to extract structural pixel information. Yuan et al. \cite{journals_tgrs_YuanWWYL11} presented an automatic road extraction method for remote sensing images based on locally excitatory globally inhibitory oscillator networks. A multistage framework to extract roads from the high-resolution multispectral satellite image was introduced by Das et al. \cite{journals_tgrs_DasMV11}. In this method, probabilistic support vector machines and salient features were used.
\par Recently, a convolutional neural network based algorithm was introduced to learn features from noisy labels in \cite{conf_eccv_AlvarezGLL12}, in which the training labels were generated by applying an algorithm trained on a general image dataset. Mnih et al. \cite{conf_eccv_MnihH10} proposed a deep neural network method, which is based on restricted boltzmann machines (RBMs), to extract urban road network from high-resolution images. In this method, unsupervised pretraining and supervised post-processing were introduced to improve the performance of the road detector substantially. A higher-order conditional random field (CRF) model was applied for road network extraction by Wegner et al. \cite{conf_cvpr_WegnerMS13}, in which the road prior was represented by higher-order cliques that connect sets of superpixels along straight line segments, and the road likelihood was amplified for thin chains of superpixels.
\par Most popular and successful road centerline extraction methods consist of one or two processing steps: 1) classification and 2) centerline extraction. Zhu et al. \cite{journals_ijrs_cZhu05} proposed a road centerline extraction approach, which is based on the binary-greyscale mathematical morphology and a line segment match algorithm. An accurate centerline detection and line width estimation method via radon transform was introduced in \cite{journals_tip_ZhangC07}. Gamba et al. \cite{journals_igrsl_PGamba06} extracted the road centerline with the help of adaptive filtering and perceptual grouping. A novel road centerline extraction method was proposed in \cite{journals_ijrs_Huangxin09} by integrating mutiscale spectral-structural features, support vector machines (SVMs) and morphological thinning algorithm.
\par In recent years, Poullis and You \cite{journals_isprs_CPSY10} proposed a novel vision-based system for automatic road centerline extraction. This system integrated Gabor filtering, tensor voting and global optimization using graph-cuts into a unified framework. A novel system \cite{journals_tgrs_UnsalanS12} was introduced to extract road centerline form high resolution images, in which probabilistic road center detection, road shape extraction, and graph-theory-based road network formation are utilized. Chaudhuri et al. \cite{journals_jstars_DCNKAS12} presented a semi-automatic road centerline extraction algorithm. To achieve this, directional morphological enhancement and directional segmentation were used to extract the road area, then thinning method was applied to extract the road network. An automatic road centerline extraction method was introduced by Miao et al. \cite{journals_lgrs_MiaoSZW13}, in which potential road segments were obtained based on shape features and spectral features, followed by multivariate adaptive regression splines to extract road centerlines.
\par Shi et al. \cite{journals_lgrs_ShiMWZ14} presented a two-step method for urban road extraction. First, spectral-spatial classification and shape features were employed to obtain road segmentation results. Then morphological thinning algorithm was used to extract centerlines. An integrated urban main-road centerline detection method was introduced in \cite{journals_tgrs_ShiMD14}. Road extraction result was achieved by fusing spectral-spatial classification and local Geary's C method. Then, to extract smooth road centerlines, local linear kernel smoothing regression algorithm was introduced. It relieves the shortcoming of thinning algorithm, while it can't work well in the intersection areas. Hu et al. \cite{journals_tgrs_HuLSZZ14} proposed a three-step road centerline extraction method. First, adaptive mean shift was used to detect road center points. Then, the salient linear features were enhanced via stick tensor voting. Finally, a weighted Hough transform was applied to extract the arc primitives of the road centerlines. Sironi et al. \cite{conf_cvpr_SironiLF14} provided a new idea by setting the road centerline extraction task as a regression problem, which learns a scale space distance transform from the high-resolution image.
\begin{figure*}[t]
\centerline{\includegraphics[width=16.7cm]{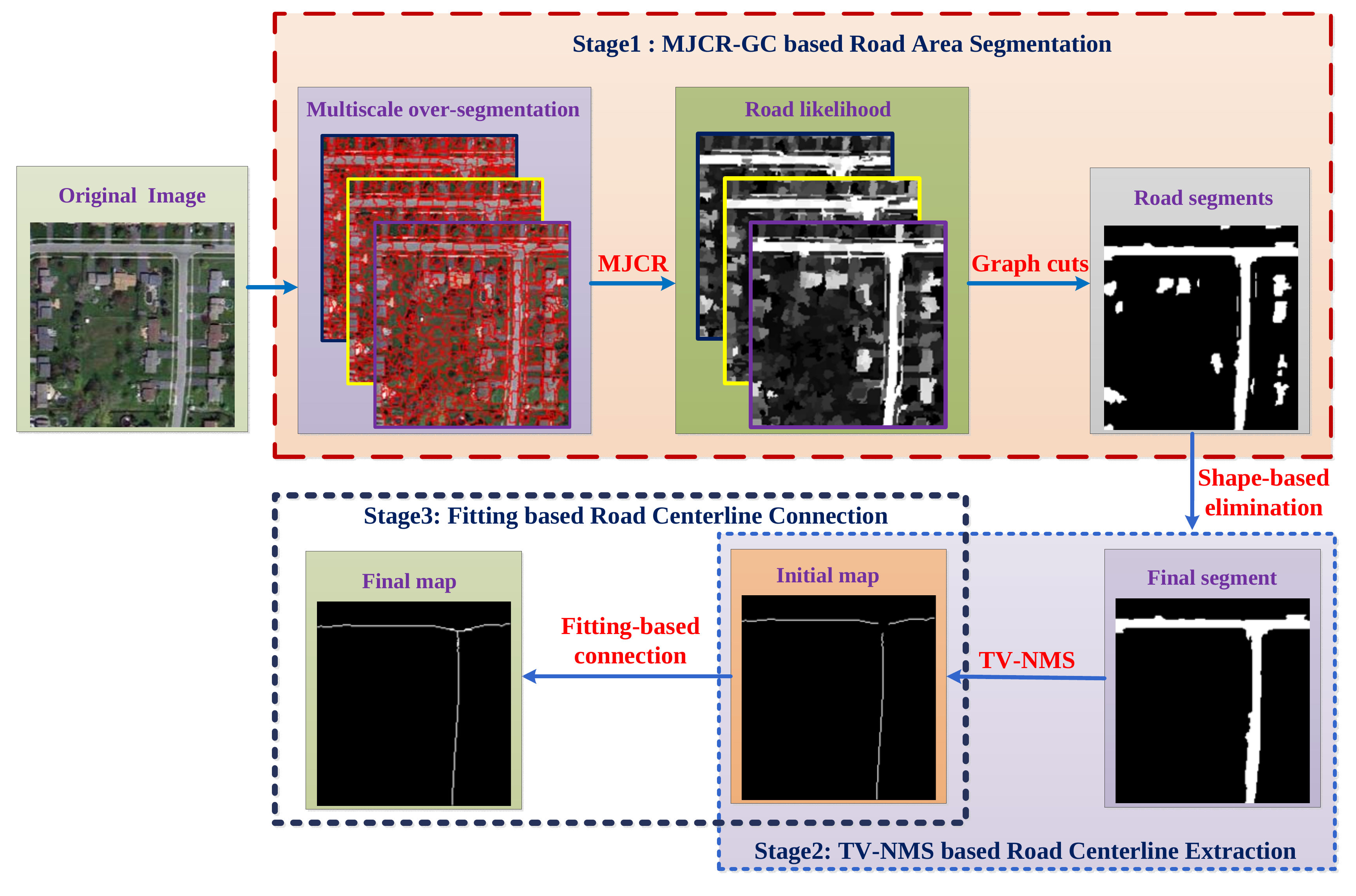}}
\caption{Flowchart of our method. It contains three stages: road area segmentation, road centerline extraction, and road centerline connection. In the $1^{\textrm{st}}$ stage, multiscale joint collaborative representation (MJCR) and graph cuts (GC) are applied to obtain the road segmentation result, then a shape based elimination algorithm is introduced to distinguish the road segments and road-like segments. For the $2^{\textrm{nd}}$ stage, tensor voting (TV), non-maximum suppression (NMS) are proposed to extract smooth and correct road centerlines. A fitting based connection algorithm is utilized to complete the road centerline network in the $3^{\textrm{rd}}$ stage.}
\label{flow_chart}
\end{figure*}
\section{Road Segmentation and Centerline Extraction}
\label{Section3}
\par In this section, we propose a novel urban road centerline extraction method, which consists of three stages: road area segmentation, road centerline extraction and road centerline connection. The flow chart is shown in Fig. \ref{flow_chart}. Specifically, in the $1^{\textrm{st}}$ stage, MJCR and GC are combined to gain an accurate road area segmentation result. Then, to eliminate those road-like segments from the segmentation result, a novel shape based elimination algorithm is introduced. To obtain a smooth and correct road centerline network, the $2^{\textrm{nd}}$ stage provides the tensor voting (TV) and non-maximum suppression (NMS) based road centerline extraction method. In the $3^{\textrm{rd}}$ stage, to obtain a complete road centerline network, a fitting based connection algorithm is utilized around the road intersections.
\subsection{\bf MJCR-GC based Road Area Segmentation}
\par In this subsection, we propose two techniques, i.e. multiscale joint collaborative representation and graph cuts (MJCR-GC), to classify the image into road and non-road areas. Specifically, there are three stages. First, we present a novel object-based spectral, structural and contextual (SSC) method to extract features. Second, a MJCR-GC based framework is proposed to obtain an accurate road area segmentation result. Third, to remove the non-road objects (i.e. houses and parking lots) from the road class, a new shape-based elimination algorithm will be employed.
\subsubsection{\bf Object-based SSC feature extraction}
In the VHR remote sensing image, the road segments tend to be elongated and under the occlusions of cars and trees etc. Based on this observation, we employ an object-oriented SSC feature to achieve the following two goals: (1) to reduce the side influence of occlusions and the spectral variations; (2) to extract the geometric characteristics of road segments. The basic assumption behind the object-based algorithm is that spatially adjacent pixels are prone to be grouped into spectrally homogeneous objects. Specifically, the simple linear iterative clustering (SLIC) \cite{journals_pami_AchantaSSLFS12} is employed to obtain the multiscale over-segmentation for VHR remote sensing images. SLIC adapts a local k-means clustering approach according to their color similarity and proximity in the image plane. There are three advantages to apply the SLIC for the over-segmentation task. SLIC is not only more efficient, but also achieving more satisfactory over-segmentation results compared with the state-of-the-art methods, for example graph-based algorithms \cite{journals_pami_ShiM00, journals_ijcv_FelzenszwalbH04} and gradient-ascent based algorithms \cite{journals_pami_ComaniciuM02, conf_eccv_VedaldiS08} etc. Besides, there is only one parameter specifies the number of superpixels, which makes it extremely easy to use.
\par We obtain multiscale segmentations with different superpixel numbers. Intuitively, superpixel number controls the segment size. A large superpixel number will averagely result in a small object size, and vice versa. As Fig. \ref{flow_chart} shows, we get three different over-segmentation results for each image.
\begin{figure}[t]
\centerline{\includegraphics[width=8.3cm]{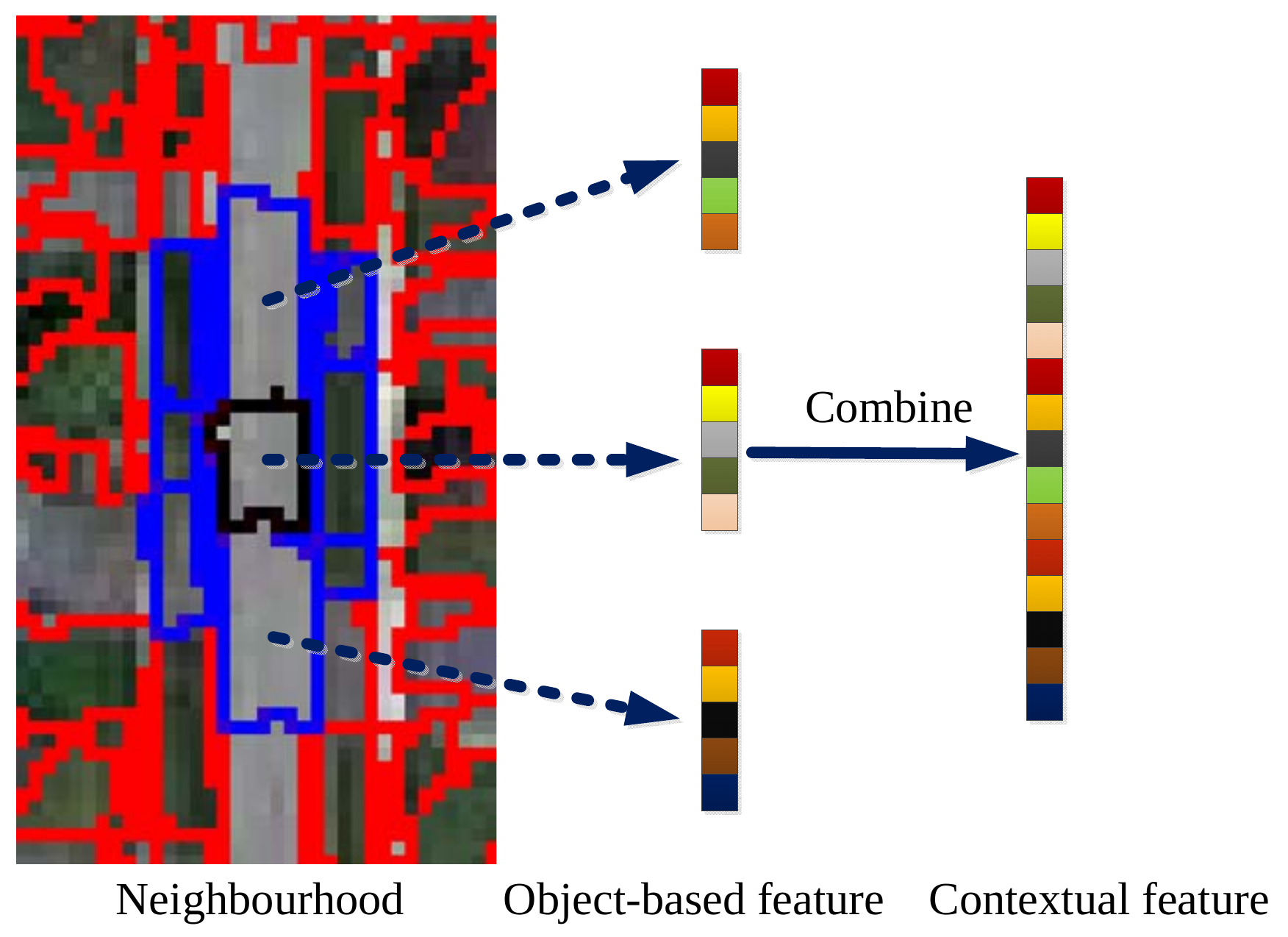}}
\caption{The illustration of how to extract the spectral, structural and contextual feature for the superpixel with dark border.}
\label{neighborhood_cut}
\end{figure}
\par For each over-segmentation result, we define the spectral attribute of an object as the average spectral value within this object. For the $i$-th object, the spectral attribute (SA) \cite{journals_ijrs_Huangxin09} is given by
\begin{equation}
\label{Eq:SA}
\begin{aligned}
\textrm{SA}_{i}&=\{\textrm{SA}_{i}^{r},\textrm{SA}_{i}^{g},\textrm{SA}_{i}^{b}\} \\
\underset{m\in \{r,g,b\}} {\textrm{SA}_{i}^{m}}&=\frac{1}{N}_{i}\underset{p\in \textrm{obj}_{i}}{\sum}\textrm{SP}^{m}(p),
\end{aligned}
\end{equation}
where $p$ is a pixel in the $i$-th object $(1\leq p \leq N_{i})$ and $\textrm{SP}^{m}(p)$ is the spectral value of pixel $p$ in band $m$; The superscripts $r$, $g$ and $b$ represent the RGB channels, respectively. To exploit the structural attributes for each object, we utilize the {\it shape index} (SI) \cite{journals_ijrs_Huangxin09, journals_pers_MingjunSong04} and {\it aspect ratio} (AR) as follows
\begin{equation}
\label{Eq:SI}
\textrm{SI}_{i}=\frac{\textrm{Per}_{i}}{4\sqrt{\textrm{Area}_{i}}},
\end{equation}
\begin{equation}
\label{Eq:AR}
\textrm{AR}_{i}=\frac{\textrm{Length}_{i}}{\textrm{Width}_{i}},
\end{equation}
where $\textrm{Per}_{i}$ and $\textrm{Area}_{i}$ denote the perimeter and area of the $i$-th object, respectively; $\textrm{Length}_{i}$ and $\textrm{Width}_{i}$ are the length and width of the minimum bounding box surrounding the $i$-th object.
\par The SI measures the smoothness of the object border and the AR describes the length-width ratio. Intuitively road regions tend to be elongated ribbons with large perimeters and small areas. Therefore, the SI and AR of road regions tend to be large.
\par The aforementioned spectral and structural attributes for each object can be combined as a hybrid feature (HF), which is defined as
\begin{equation}
\label{Eq:HF}
\textrm{HF}_{i}^{s}=\{\textrm{SA}_{i}^{s},\textrm{SI}_{i}^{s},\textrm{AR}_{i}^{s}\},
\end{equation}
where $\textrm{HF}_{i}^{s}$ is the hybrid feature for the $i$-th object at the $s$-th scale.
\par To enhance the discriminative power of each object, the spatially adjacent objects should be considered. In the proposed method, co-occurrence among objects are used as the high-level contextual feature. Intuitively, there are at least two road segments (i.e. left and right, or up and down) in the neighborhood of one road segment. This is the prior information for humans to recognize the road areas under the occlusions of trees and shadows. Our SSC feature employs this recognition scheme. Fig.~\ref{neighborhood_cut} illustrates how to extract the SSC feature for one center segment. There are three steps. First, we find out the neighboring segments (i.e. the blue border regions) for the center segment (i.e. the dark border region). Second, we rank the neighboring segments according to the their similarity values to the center segment, where we use the Euclidean distance as the similarity measure. Third, the HFs of both center segment and its top two neighboring segments are stacked as the contextual feature for the center segment.
\subsubsection{\bf MJCR-GC based road area segmentation}
In this subsection, MJCR and GC are combined to obtain an accurate road area segmentation result. Specifically, SSC feature is utilized by MJCR to gain the road probability of each segment. Then, to enhance the label consistency among the neighboring pixels, GC is introduced to incorporate the spatial information.
\paragraph{\bf MJCR based road likelihood} Recently, sparse representation classification (SRC) \cite{journals_pami_WrightYGSM09} has been proposed for face recognition. SRC represents a testing sample by a sparse linear combination of training samples with $\ell_{1}$-norm constraint. In the remote sensing imagery, Chen et al. \cite{journals_tgrs_ChenNT11} applied a sparse framework for the hyperspectral image classification. A similar approach to SRC is the collaborative representation classification (CRC) \cite{journals_tgrs_LiTPF14, journals_lgrs_LiDZH15}. CRC also represents a testing sample with the linear combination of training samples. However, contrary to the $\ell_{1}$-norm regularization in SRC, CRC employs an $\ell_{2}$-norm regularization. It provides competitive accuracy while with significantly lower computational complexity.
\par Consider a dataset with $n$ training samples $\mathbf{X}=\{\mathbf{x}_{i}\}_{i=1}^{n}$, where $\mathbf{x}_{i}\in{\mathbb{R}}^{d}$ and $d$ is the feature dimensionality. Let $y_{i}\in\{1,2,...C\}$ be the class label, where $C$ is the number of classes. For the road extraction task, we set $C=2$. Let $n_{1}$ and $n_{2}$ be the number of training samples for road class and non-road class, respectively.
\par For a testing sample $\widetilde{\mathbf{x}}$, the corresponding combination coefficient $\bm{\alpha}$, based on all the training samples, could be obtained via
\begin{equation}
\label{Eq:JRC}
\bm{\alpha}^*=\underset{\bm{\alpha}}{\text{arg min}} \ ||\widetilde{\mathbf{x}}-\mathbf{X}\bm{\alpha}||_{2}^{2}+\lambda \, ||\Gamma_{\mathbf{X},\widetilde{\mathbf{x}}}\bm{\alpha}||_{2}^{2},
\end{equation}
where $\Gamma_{\mathbf{X},\widetilde{\mathbf{x}}}$ is a biasing Tikhonov matrix between the test sample and all the training samples; $\lambda$ is a global regularization parameter that balances the representation loss and the regularization term. Note that $\bm{\alpha}^*$ is the optimal representation vector of $\bm{\alpha}$ with $n\times 1$ elements. Specifically, $\Gamma_{\mathbf{X},\widetilde{\mathbf{x}}}\in{\mathbb{R}}^{n\times n}$ is designed in the following form:
\begin{equation}
\begin{aligned}
\label{Eq:Gamma}
\Gamma_{\mathbf{X},\widetilde{\mathbf{x}}}=\left[\begin{array}{ccc}
||\widetilde{\mathbf{x}}-\mathbf{x}_{1}||_{2}&\cdots&0 \\
\vdots&\ddots&\vdots \\
0&\cdots&||\widetilde{\mathbf{x}}-\mathbf{x}_{n}||_{2}
\end{array}\right].
\end{aligned}
\end{equation}
$\Gamma_{\mathbf{X},\widetilde{\mathbf{x}}}$ is a diagonal matrix, whose diag value measures the discrepancy between a certain training sample and the testing sample. Intuitively, if the testing sample belongs to the road class, the discrepancies between the testing sample and those road-class training samples are small, while inversely the discrepancies are large for the non-road-class training samples. Given a large regularization parameter $\lambda$, to achieve a minimum objective in Eq. (\ref{Eq:JRC}), the road-class testing sample are more likely to be represented by road-class samples rather than non-road-class samples, thus $\bm{\alpha}^*$ tend to be sparse.
\par The representation coefficient $\bm{\alpha}^*$ can be estimated in a closed-form solution as
\begin{equation}
\label{Eq:alphastar}
\bm{\alpha}^*=\Big(\mathbf{X}^{T}\mathbf{X}+\lambda \Gamma^{T}_{\mathbf{X},\widetilde{\mathbf{x}}}\Gamma_{\mathbf{X},\widetilde{\mathbf{x}}}\Big)^{-1}\mathbf{X}^{T}\widetilde{\mathbf{x}}.
\end{equation}
\par After that, the training samples $\mathbf{X}$ are partitioned as road-class samples $\mathbf{X}_{1}$ and non-road samples $\mathbf{X}_{2}$, and the coefficient vector $\bm{\alpha}^*$ are partitioned as $\bm{\alpha}^*_{1}$ and $\bm{\alpha}^*_{2}$ accordingly. The residual between the approximation and the testing sample can be defined as
\begin{equation}
\label{Eq:residual}
\underset{l\in\{1,2\}}{R_{l}(\widetilde{\mathbf{x}})}=||\mathbf{X}_{l}\bm{\alpha}^*_{l}-\widetilde{\mathbf{x}}||_{2}^{2}.
\end{equation}
\par In this research, we get the class likelihood rather than the class label of each sample. Therefore, we define the road-class likelihood as
\begin{equation}
\label{Eq:road_likelihood}
p_{r}(\widetilde{\mathbf{x}})=\frac{{R_{2}(\widetilde{\mathbf{x}})}}{{R_{1}(\widetilde{\mathbf{x}})}+{R_{2}(\widetilde{\mathbf{x}})}},
\end{equation}
and the non-road-class likelihood as $p_{nr}(\widetilde{\mathbf{x}})=1-p_{r}(\widetilde{\mathbf{x}})$.
\par After obtaining the road likelihood for each object at a certain scale, all the pixels in the same object are given an identity likelihood value as the object. Then, we fuse all road likelihood maps from the three different scales into an integrated one. Finally, the road likelihood of each pixel $x_{i}$ can be defined as
\begin{equation}
\label{Eq:fusion_likelihood}
p_{r}(x_{i})=\underset{s\in\{1,2,3\}}{\textrm{max}}p_{r}^{s}(x_{i}).
\end{equation}
We also tried other fusion strategies, such as {\it mean, median} and {\it min}. Their results are inferior to the {\it max} rule. Thus we use the {\it max}-fusion rule in our experiments.
\paragraph{\bf GC based road area segmentation}
In the remote sensing image, some road areas are under conditions of spectral variability as well as occlusions of trees and cars. To relieve the side effect of these conditions and get a coherence road extraction result, graph cuts (GC) algorithm is used for the road extraction task.
\par Given an image $I$, the GC algorithm constructs an undirected graph $\mathcal{G}=\{\mathcal{V},\mathcal{E}\}$, where $\mathcal{V}$ denotes the pixel set in the image and $\mathcal{E}$ represents the set of undirected graph edges between neighbouring pixels \cite{conf_icip_ChengWGZP14}. For the road extraction task, we define the label ``1'' for the road class and ``0'' for the non-road class. GC tries to minimize the following objective
\begin{equation}
\label{Eq:GC}
C(\mathcal{L})=C_{r}(\mathcal{L})+\alpha C_{b}(\mathcal{L}),
\end{equation}
where $\mathcal{L}$ is a labeling set, $C_{r}(\mathcal{L})$ and $C_{b}(\mathcal{L})$ denote the regional term and boundary term, respectively; $\alpha$ is a trade-off parameter balancing the two terms. In the road extraction problem, the regional term $C_{r}(\mathcal{L})$ defines the individual penalty to classify each pixel into the road class. The boundary term $C_{b}(\mathcal{L})$ describes the coherence between spatially neighboring pixels.
\par We have obtained the road likelihood map via aforementioned MJCR. The regional term is defined as
\begin{equation}
\label{Eq:regional}
C_{r}(\mathcal{L})=\underset{i\in I}{\sum}-\textrm{log}\big(p_{r}(x_{i})\big),
\end{equation}
where $p_{r}(x_{i})$ is the road likelihood probability of pixel $x_{i}$. Intuitively, the spatially adjacent pixels tend to belong to the same class, thus the boundary term is defined to measure the label discontinuity among neighboring pixels, which is defined as
\begin{equation}
\label{Eq:boundary}
C_{b}(\mathcal{L})=\mathop\sum_{{i,j}\in \mathcal{N}} m({L_{x_{i}},L_{x_{j}}})\cdot \frac {1} {||{\bm x_{i}}-{\bm x_{j}}||_{2}+\epsilon},
\end{equation}
where $\mathcal{N}$ denotes a standard 8-neighborhood system, which contains all the unordered pairs of neighbouring pixels. $m({L_{x_{i}},L_{x_{j}}})$ is the distance metric between the label $L_{x_{i}}$ and $L_{x_{j}}$, i.e. if $L_{x_{i}}$ and $L_{x_{j}}$ have different labels, we denote $m({L_{x_{i}},L_{x_{j}}})=1$, otherwise, we define it as $0$. ${\bm x_{i}}$ and ${\bm x_{j}}$ are the RGB feature vectors of pixel $x_{i}$ and $x_{j}$. $||\cdot||_{2}$ denotes the $\ell_2$-norm. To avoid a zero divisor, we add a small value $\epsilon$ (typically $\epsilon=0.001$) to the denominator.
\par For binary labeling problem, the objective function in Eq. (\ref{Eq:GC}) can achieve the optimal solution via the mincut/maxflow algorithm in polynomial time \cite{journal_pami_BoykovK04}. As Fig. \ref{flow_chart} shows, coherence road segmentation result can be obtained after the GC algorithm.
\par As Fig. \ref{flow_chart} shows, coherence road segmentation result can be obtained after the GC algorithm. However, it remains some road-like objects. Some strategies should be taken to remove those road-like objects.
\subsubsection{\bf Shape based elimination algorithm}
\begin{figure}[t]
\centering
\includegraphics[width=86mm]{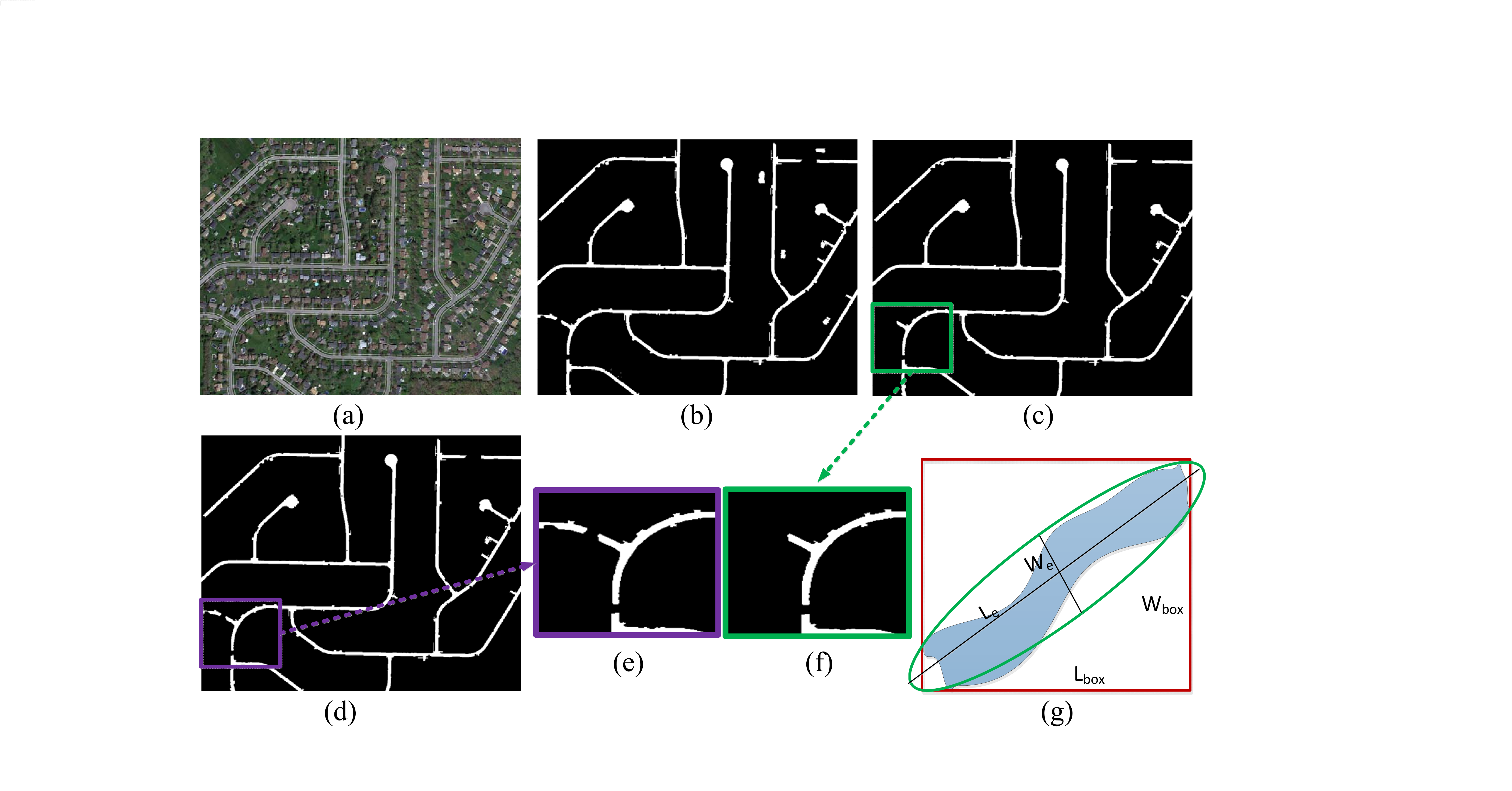}
\caption{Comparison of elimination results between our ellipse-based algorithm and the bounding box based algorithm. (a) Original image. (b) Final segmentation result. (c) Elimination result via the bounding box based algorithm. (d) Elimination result by ellipse-based algorithm. (e) and (f) are the close ups of the rectangle region in (d) and (c) respectively. (g) The illustration of the ellipse-based algorithm and the bounding box based algorithm.}
\label{Fig:road_ellipse}
\end{figure}
\par After the process of the GC based road area segmentation, there still remains some road-like objects (see Fig. \ref{Fig:road_ellipse}(\textcolor{blue}{b})). To address this issue, the road shape features should be employed to distinguish the potential road segments and road-like segments.
\par In general, roads have the following characteristics: \textbf{1)} Roads are connected and do not have small areas; In our experiments, we regard all the segments whose pixel number is less than a threshold $K$ as non-road class. \textbf{2)} Roads are elongated ribbons with long length and short width. Linear feature index (LFI) is defined to evaluate this characteristic, which can be denoted as
\begin{equation}
\label{Eq:lambda_gra}
\textrm{LFI}=\frac {\textrm{L}_\textrm{box}} {\textrm{W}_\textrm{box}}.
\end{equation}
where $\textrm{L}_\textrm{box}$ and $\textrm{W}_\textrm{box}$ are the length and width of the bounding box (the red rectangle in Fig.~\ref{Fig:road_ellipse}(\textcolor{blue}{g})), respectively. Intuitively, road segments have large LFI values. However, the bounding box based algorithm may be failed as shown in Figs.~\ref{Fig:road_ellipse}(\textcolor{blue}{c}) and \ref{Fig:road_ellipse}(\textcolor{blue}{f}), some road segments (the sketch map in Fig.~\ref{Fig:road_ellipse}(\textcolor{blue}{g})) may have small LFI values.
\par To overcome the shortcoming above, we propose a new ellipse-based road elimination algorithm (see the green ellipse in Fig.~\ref{Fig:road_ellipse}(\textcolor{blue}{g})). We use an ellipse to match each remaining segment after the area constraint. We define the new LFI as
\begin{equation}
\label{Eq:lambda_gra}
\textrm{LFI}_\textrm{e}=\frac {\textrm{L}_\textrm{e}} {\textrm{W}_\textrm{e}},
\end{equation}
where $\textrm{L}_\textrm{e}$ and $\textrm{W}_\textrm{e}$ are the major axis and minor axis of ellipse, respectively. In our experiments, we set the $\textrm{LFI}_\textrm{e}$ as $3$. Comparing Figs.~\ref{Fig:road_ellipse}(\textcolor{blue}{c}), \ref{Fig:road_ellipse}(\textcolor{blue}{d}), \ref{Fig:road_ellipse}(\textcolor{blue}{e}) and \ref{Fig:road_ellipse}(\textcolor{blue}{f}), the real road segment is removed in Figs.~\ref{Fig:road_ellipse}(\textcolor{blue}{c}) and \ref{Fig:road_ellipse}(\textcolor{blue}{f}) via bounding box constraint, while our ellipse-based elimination algorithm keeps it remained in Figs.~\ref{Fig:road_ellipse}(\textcolor{blue}{d}) and \ref{Fig:road_ellipse}(\textcolor{blue}{e}). Thus, it demonstrates that our proposed algorithm is more effective than the bounding box method.
\begin{figure}[t]
\centerline{\includegraphics[width=8.5cm]{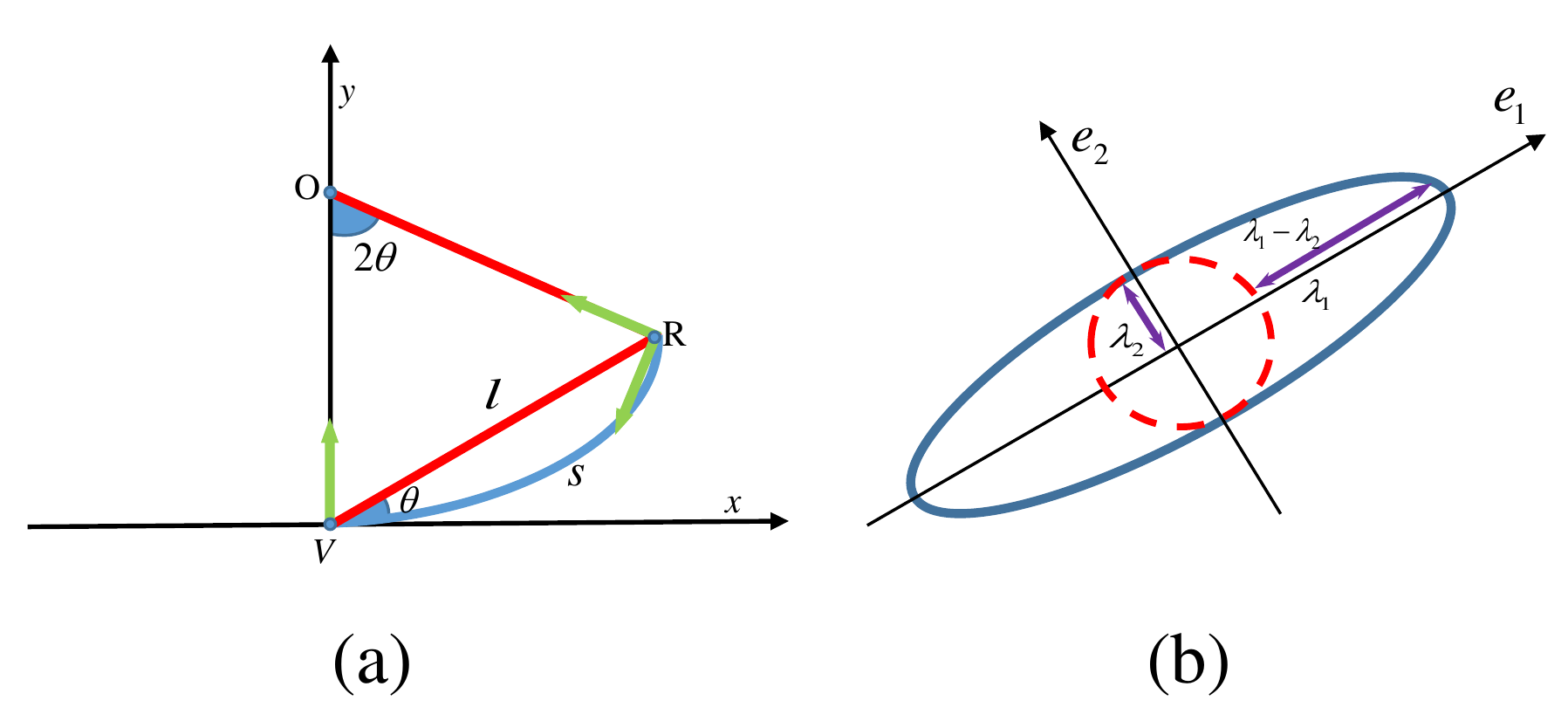}}
\caption{Illustration of tensor voting algorithm. (a) voting process. The voter (V) casts its votes to the receiver (V). (b) The sketch map of different saliency values corresponding to different eigenvectors for a tensor.}
\label{tensor_voting}
\end{figure}
\subsection{\bf TV-NMS based Road Centerline Extraction}
\par After the above processes, there are still two issues with the gained road segmentation results. Due to the image noise and occlusions caused by trees and cars, the extracted road area always have an unsatisfied property, that is, discontinuities. Besides, for some real problems, we only consider whether there exists a road, and do not care about the width of the road. To relieve the side effect of discontinuity and to be more suitable for application, road centerline extraction has been an active research. The commonly used road centerline extraction algorithm is morphological thinning, which is fast and easy to implement. However, thinning based road centerline extraction algorithm always produces many spurs and bring in many false positives, which reduce the smoothness and correctness of road network. To solve this problem, TV-NMS is introduced to extract road centerlines, which can produce smooth centerlines and complete the discontinuous road centerlines.

\par Tensor voting (TV) was originally proposed by Guy and Medioni \cite{journals_ijcv_GuyM96}, and later presented a lot of improved versions \cite{series_synthesis_2006Mordohai, conf_eccv_FrankenARFR06}. It is a technique for perceptual grouping and robust extraction of lines and curves from images. TV aims at making the local feature measurements more robust by taking the neighboring measurements into consideration. In 2-D tensor voting, local image feature are encoded into a tensor filed, in which each tensor $\mathbf{S}$ at a certain location can be denoted as a symmetric, positive semi-definite matrix representation as
\begin{equation}
\begin{aligned}
\label{Eq:Tensor}
\mathbf{S}=\left[\begin{array}{cc}
s_{xx}&s_{xy} \\
s_{xy}&s_{yy} \\
\end{array}\right]=\lambda_{1}\mathbf{e}_{1}{\mathbf{e}_{1}}^{T}+\lambda_{2}\mathbf{e}_{2}{\mathbf{e}_{2}}^{T},
\end{aligned}
\end{equation}
where $\lambda_{1}$ and $\lambda_{2}$ are the nonnegative eigenvalues $(\lambda_{1} \geq\lambda_{2}\geq0)$ of the matrix $\mathbf{S}$; $\mathbf{e}_{1}$ and $\mathbf{e}_{2}$ are their corresponding eigenvectors. As Fig. \ref{tensor_voting}(\textcolor{blue}{b}) shows, the graphical illustration of the tensor $\mathbf{S}$ is an ellipse. The tensor $\mathbf{S}$ can be decomposed as
\begin{equation}
\label{Eq:Tensor2}
\mathbf{S}=(\lambda_{1}-\lambda_{2})\mathbf{e}_{1}{\mathbf{e}_{1}}^{T}+\lambda_{2}\Big(\mathbf{e}_{1}{\mathbf{e}_{1}}^{T}+\mathbf{e}_{2}{\mathbf{e}_{2}}^{T}\Big),
\end{equation}
\begin{figure}[t]
\centerline{\includegraphics[width=8.5cm]{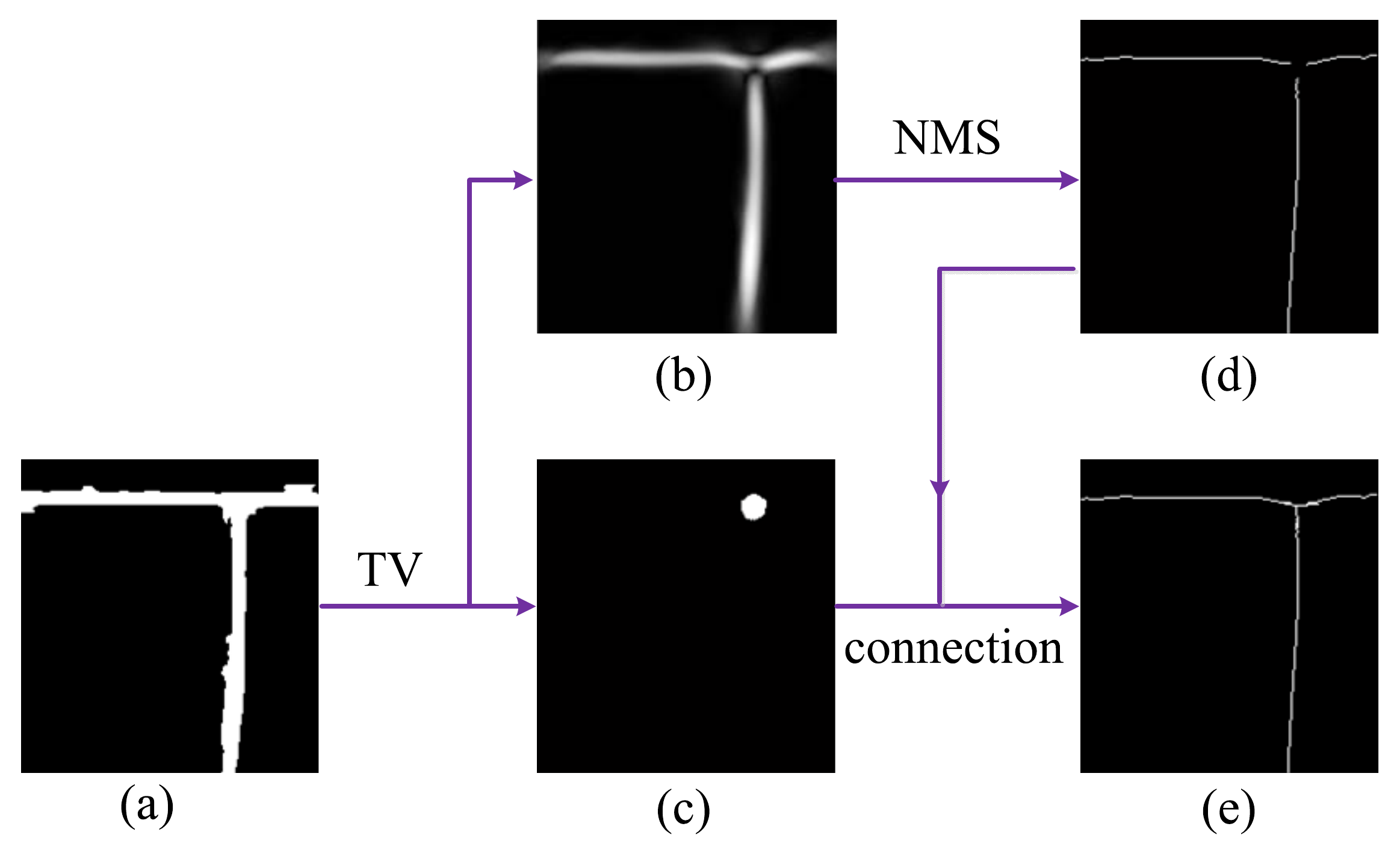}}
\caption{The flowchart of the proposed centerline extraction method. (a) Road segmentation result. (b) Stick saliency map after tensor voting. (c) Junction map after tensor voting. (d) Incomplete centerline map after non-maximum suppression (NMS). (e) Complete centerline map after the fitting based centerline connection. }
\label{TV_NMS}
\end{figure}
where $\mathbf{e}_{1}{\mathbf{e}_{1}}^{T}$ corresponds to a degenerate elongated ellipsoid, termed the {\it stick tensor}; $(\mathbf{e}_{1}{\mathbf{e}_{1}}^{T}+ \mathbf{e}_{2}{\mathbf{e}_{2}}^{T})$ corresponds to a circular disk, termed the {\it ball tensor}; $(\lambda_{1}-\lambda_{2})$ and $\lambda_{2}$ are the saliency values of the stick tensor and ball tensor respectively. The spectrum theorem \cite{books_SPCV} states that any 2-D tensor can be decomposed into the linear combination of stick tensor and ball tensor.
\par In the voting process, as Fig. \ref{tensor_voting}(\textcolor{blue}{a}) shows, assume point V is the {\it voter} and point R is the {\it receiver}, $\odot$O is the osculating circle passing V and R. The second order vote is a stick tensor and has a normal lying along the radius of the osculating circle at R \cite{books_ETCV}. According to the Gestalt
principles, the magnitude of the vote should be a function of proximity and smooth continuation. Thus the saliency decay function can be defined as
\begin{equation}
\label{Eq:decay}
DF(s,\kappa,\sigma)=e^{-\big(\frac{s^{2}+c\kappa^{2}}{\sigma^{2}}\big)},
\end{equation}
where $s=\frac{l\theta}{2\textrm{sin}{\theta}}$ is the arc length of VR; $l$ is the distance between the voter and receiver; $\theta$ is the angle between the tangent of the osculating circle at the voter and the line going through the voter and receiver (see Fig. \ref{tensor_voting}(\textcolor{blue}{a})); $\kappa=\frac{2\textrm{sin}\theta}{l}$ is the curvature, and $c$ controls the degree of the decay with curvature; $\sigma$ is the scale of voting, which determines the effective neighborhood size. In the experiment, the parameter $c$ is a function of the scale \cite{books_ETCV} as
\begin{equation}
\label{Eq:cvalue}
c=\frac{-16(\sigma-1)\times \textrm{log}(0.1)}{{\pi}^{2}}.
\end{equation}

In this research, the voting process includes two kinds of voting: {\it sparse voting} and {\it dense voting}. The sparse voting describes a pass of voting where votes are cast to locations that contain tokens only. The dense voting denotes a pass of voting from the tokens to all the locations within the neighborhood regardless of the presence of the tokens. Then the receivers accumulate the votes cast to them by tensor addition.
\par After tensor voting, according to the saliency value, the point feature can be classified into three types (i.e. {\it curve point}, {\it junction point} and {\it outlier}) \cite{journals_ijcv_GuyM96} as follows,
\begin{equation}
\label{MC_point_feature}
\begin{cases}
\textrm{curve point} & \text{if}\ (\lambda_{1}-\lambda_{2})>\lambda_{2} \\
\textrm{junction point} & \text{if}\ \lambda_{1}\thickapprox \lambda_{2}>0 \\
\textrm{outlier} & \text{if}\ \lambda_{1}\thickapprox 0, \lambda_{2}\thickapprox 0
\end{cases}.
\end{equation}
 According to the above rules, we get the junction map (see Fig. \ref{TV_NMS}(\textcolor{blue}{c})) and stick saliency map (see Fig.~\ref{TV_NMS}(\textcolor{blue}{b})).
\par One great advantage of TV is that it only has one parameter (i.e. scale factor $\sigma$) to be determined by users. Actually, in most cases, the scale factor can be set according to the road width, which will be discussed later.
\par To get the consistent road centerline, Non-Maximum Suppression (NMS) is applied on the resulting stick saliency map (see Fig. \ref{TV_NMS}(\textcolor{blue}{b})). The NMS keeps only the locations that correspond to a local maximum along a line perpendicular to the local orientation within a neighborhood of width. As Fig. \ref{TV_NMS} shows, the TV-NMS method can extract smooth road centerline, while it is ineffective around the road intersection. To tackle this issue, some other measurements should be taken to complete the road centerline network.
\begin{figure}[t]
\centerline{\includegraphics[width=8.5cm]{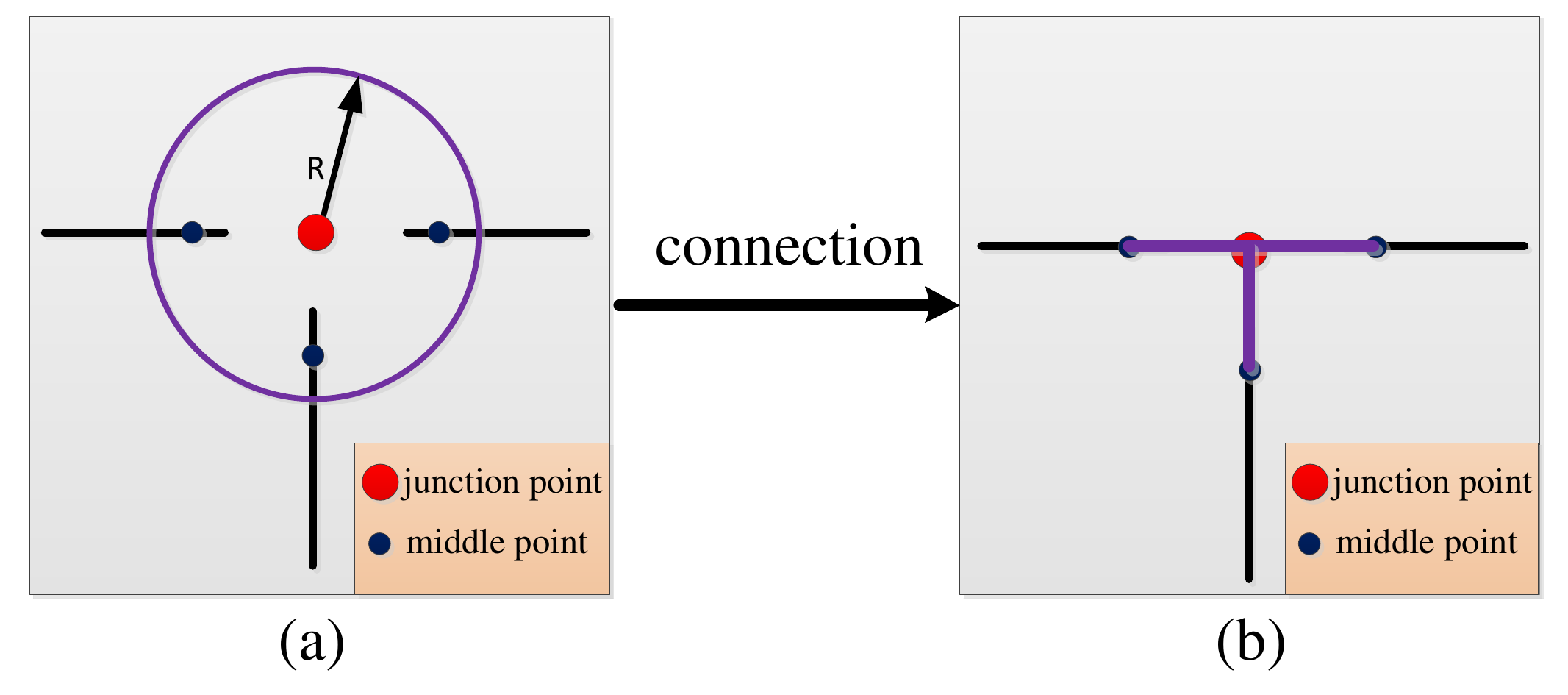}}
\caption{Illustration of the proposed fitting based connection method. (a) The sketch map of detecting the {\it middle points}. (b) The final map after connection (the purple lines).}
\label{Connection}
\end{figure}
\subsection{\bf Fitting based Road Centerline Connection}
\par TV-NMS based centerline extraction method can extract single-pixel-width road centerline well. However, it can't extract the centerlines around the road intersections properly (see the initial map in Fig.~\ref{flow_chart}). To overcome this shortcoming, a new fitting based centerline connection method is proposed.
\par We get the junction map after the TV algorithm. As Fig.~\ref{Connection}(\textcolor{blue}{a}) shows, for each junction point (the red circle), we search its local area within a certain distance defined by the radius R. For example, there are three line segments in the radius areas in Fig.~\ref{Connection}(\textcolor{blue}{a}). For each segment in the radius, we calculate its middle location of all the centerline pixels, termed {\it middle point}. Then, we link the middle point and the corresponding center point of junction area (see the purple lines in Fig.~\ref{Connection}(\textcolor{blue}{b})). In this subsection, we employ the linear fitting algorithm to connect the discontinuous road centerlines around the intersection area. It is behind the assumption that road is straight at the intersection area, which is satisfied in most cases. As Fig.~\ref{TV_NMS}(\textcolor{blue}{e}) shows, our fitting based method can link the centerlines well around the road intersection area.
\section{Experiments and Evaluation}
\label{Section4}
\par In this paper, the definition of the ``VHR" refers to the image with spatial resolution of $1$-$2$ m per pixel. The road width in VHR image is $8$-$15$ pixels. The corresponding road reference map is manually labeled.
\par To verify the effectiveness of the proposed method, extensive experiments, on the road centerline extraction from VHR remote sensing images, have been conducted on two datasets. The proposed method is also compared with other state-of-the-art methods.
\subsection{\bf Datasets description}
\par This section introduces the information of two VHR image datasets for the road extraction experiments. It should be noted that few VHR urban road datasets are publicly available. Thus we collected 30 VHR images from {\it Google Earth} and labeled the road reference map and centerline reference map by a hand drawing method. This dataset will be publicly available for research.
\par {\bf Data $\#1$:} This dataset contains $30$ VHR urban road images with a spatial resolution of $1.2 m$ per pixel. The road width is about $12$-$15$ pixels. There are at least $800\times 800$ pixels in each image. Most of the images are under the conditions of complex backgrounds and occlusions due to trees and cars. In addition, there are road-like objects (i.e. houses and parking lots) in the images. All these factors make it very challenging to obtain a satisfying road extraction results.
\par {\bf Data $\#2$:} This dataset is a publicly available road dataset%
\footnote{http://cvlab.epfl.ch/data/delin%
} provided by E. Turetken et al. \cite{conf_cvpr_TuretkenBF12}. It contains $14$ images with a spatial resolution of $1.5 m$ per pixel. The road width in this dataset is about $8$-$10$ pixels. Some images are under the conditions of complex backgrounds and occlusions of trees. We manually labeled the centerline reference map of each image in this dataset.

\subsection{\bf Compared algorithms} To verify the performance, the proposed method is compared with four related methods. The main information of all these methods are summarized as follows:
\par 1) {\bf Huang's method} (Huang): Huang et al. \cite{journals_ijrs_Huangxin09} introduce a novel road detection method based on multiscale structural features and support vector machine. Then, morphological thinning algorithm is used to extract the road centerlines.
\par 2) {\bf Miao's method} (Miao): Miao et al. \cite{journals_lgrs_MiaoSZW13} present a road extraction method based on spectral and shape features. Then, to overcome the shortcomings of morphological thinning algorithm, multivariate adaptive regression splines is introduced to get the smooth and correct road centerlines.
\par 3) {\bf Shi's method a} ($\textrm{Shi}^{{a}}$): This method \cite{journals_lgrs_ShiMWZ14} integrates the spectral-spatial classification, shape features and local Geary's C to extract road network. Then a morphological thinning algorithm is applied to extract centerlines.
\par 4) {\bf Shi's method b} ($\textrm{Shi}^{{b}}$): To get the road network, Shi et al. \cite{journals_tgrs_ShiMD14} fuse the spectral-spatial feature and homogeneous property via support vector machine. After that, a local linear kernel smoothing regression algorithm is introduced to extract the centerline. It is a state-of-the-art method for road centerline extraction.
\par 5) {\bf Proposed method with three scales} ($\textrm{Proposed}^{3}$): As Fig. \ref{flow_chart} shows, the image is oversegmented with three different numbers of superpixels, such as $8000$, $10000$ and $12000$. Then the following steps in Fig. \ref{flow_chart} are performed to get the road centerline.
\par 6) {\bf Proposed method with one scale} ($\textrm{Proposed}^{1}$): To investigate the effectiveness of multi-scale information fusion, the image is oversegmented with only one certain number of superpixels, for example $10000$. Other steps are performed as $\textrm{Proposed}^{3}$ states.
\par It should be noted that the codes of the other four comparing methods are not readily available in the Internet. We implement them according to the details described in their papers. Our implemented codes can achieve the same performance as the original paper presents. In the following experiments, we adjust the parameters to gain the best performance of the four methods for fair comparison. Among the four comparing methods, regression based centerline extraction algorithm is used in {\it Miao} and {\it $\textrm{Shi}^{b}$}, thus they can obtain smooth centerline network. Meanwhile, {\it Huang} and {\it $\textrm{Shi}^{a}$} utilize the morphological thinning algorithm to extract road centerlines, which will produce some small spurs.
\begin{figure*}[t]
\centerline{\includegraphics[width=18.3cm]{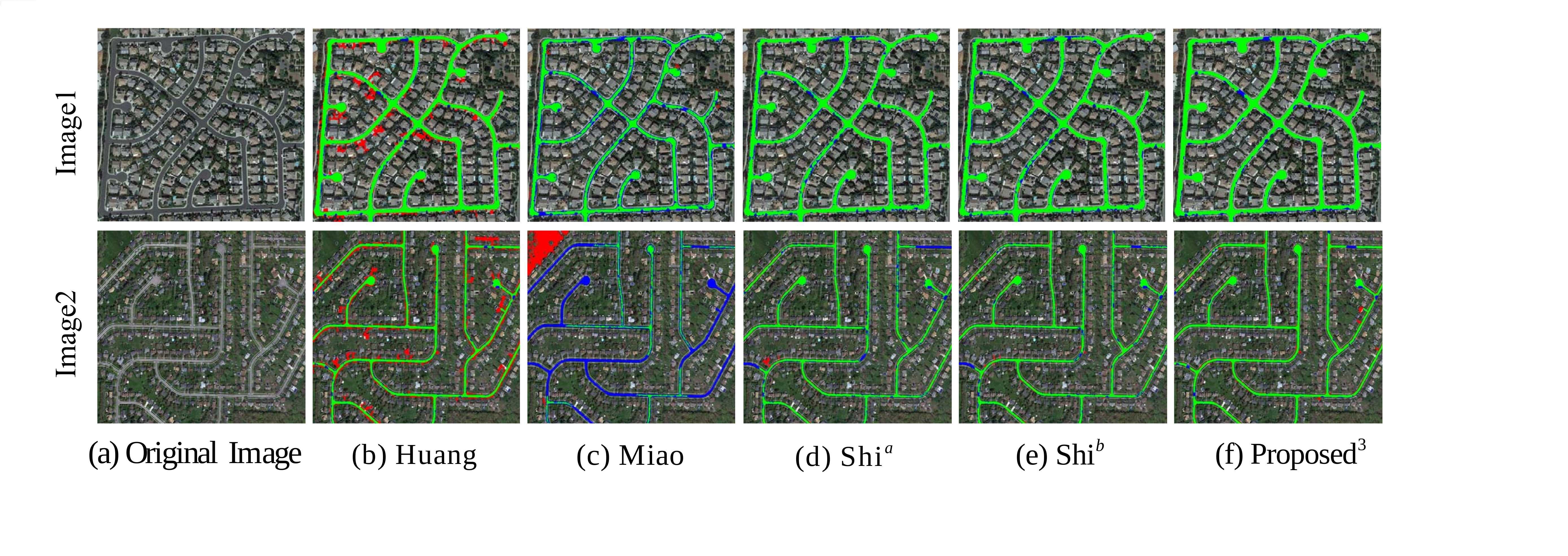}}
\caption{Visual comparisons of road area extraction results with four comparing methods. There are two rows and six columns of subfigures. Each row shows results on one dataset. From left to right: (a) original image, (b) result of Huang \cite{journals_ijrs_Huangxin09}, (c) result of Miao \cite{journals_lgrs_MiaoSZW13}, (d) result of $\textrm{Shi}^{a}$ \cite{journals_lgrs_ShiMWZ14}, (e) result of $\textrm{Shi}^{b}$ \cite{journals_tgrs_ShiMD14}, (f) result of $\textrm{Proposed}^{3}$. From (b) to (f), {\color{green}{true positive (TP)}} is marked in green, {\color{red}{false positive (FP)}} in red, {\color{blue}{false negative (FN)}} in blue. (Best viewed in color)}
\label{classification_compare}
\end{figure*}
\subsection{\bf  Evaluation metrics}
\par To comprehensively evaluate the proposed method, comparison experiments are conducted in two aspects: the comparison of road area extraction result and the comparison of the road centerline extraction result. {\it Completeness} (COM), {\it Correctness} (COR) and {\it Quality} (Q) \cite{Heipke97} are employed as evaluation metrics to measure the both experiments as follows.
\begin{equation}
\label{Eq:com}
\textrm{COM}=\frac{\textrm{TP}}{\textrm{TP}+\textrm{FN}},
\end{equation}
\begin{equation}
\label{Eq:cor}
\textrm{COR}=\frac{\textrm{TP}}{\textrm{TP}+\textrm{FP}},
\end{equation}
\begin{equation}
\label{Eq:q}
\textrm{Q}=\frac{\textrm{TP}}{\textrm{TP}+\textrm{FN}+\textrm{FP}}.
\end{equation}
where TP, FP and FN are the true positive, false positive and false negative, respectively.
\par For the evaluation of road area extraction result, we compare the extracted road map with reference map in the corresponding locations.
\begin{table}[t]
\caption{Quantitative comparisons with state-of-the-art methods in two datasets, where the {\color{red}{\textbf{red values}}} marked in bold are the best and the bold {\color{blue}{\textbf{blue values}}} are the second best. (Best viewed in color)}
\centerline{\includegraphics[width=11cm]{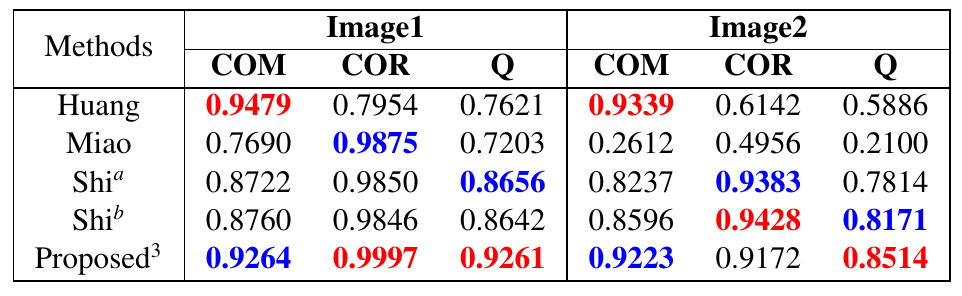}}
\label{table_classification}
\end{table}
\par Due to the deviation between the manually labeled centerline and the real centerline, the comparison of centerline result is carried out by matching the extracted centerline map to the reference map using the so-called ``buffer method", in which every proportion of the network within a given buffer width $\rho$ from the other is considered as matched \cite{BWessel03}. That is a predicted centerline point is considered to be a true positive if it is at most $\rho$ distant from a reference centerline point.
\begin{table*}[t]
\caption{Quantitative comparisons with state-of-the-art methods in Data $\#1$, where the {\color{red}{\textbf{red values}}} marked in bold are the best and the bold {\color{blue}{\textbf{blue values}}} are the second best. (Best viewed in color)
\label{centerline_table1}}
\centerline{\includegraphics[width=17cm]{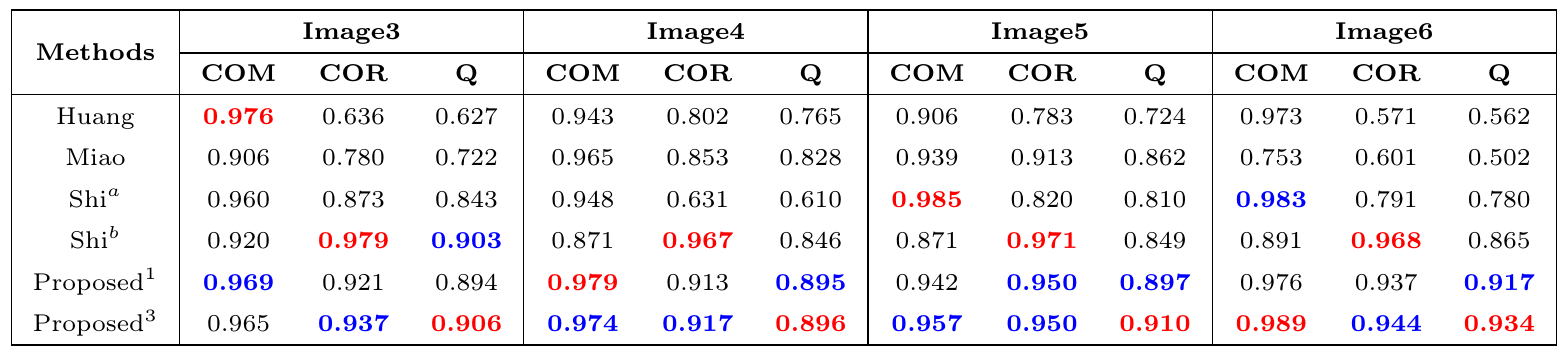}}
\end{table*}
\begin{figure*}[t]
\centerline{\includegraphics[width=17cm]{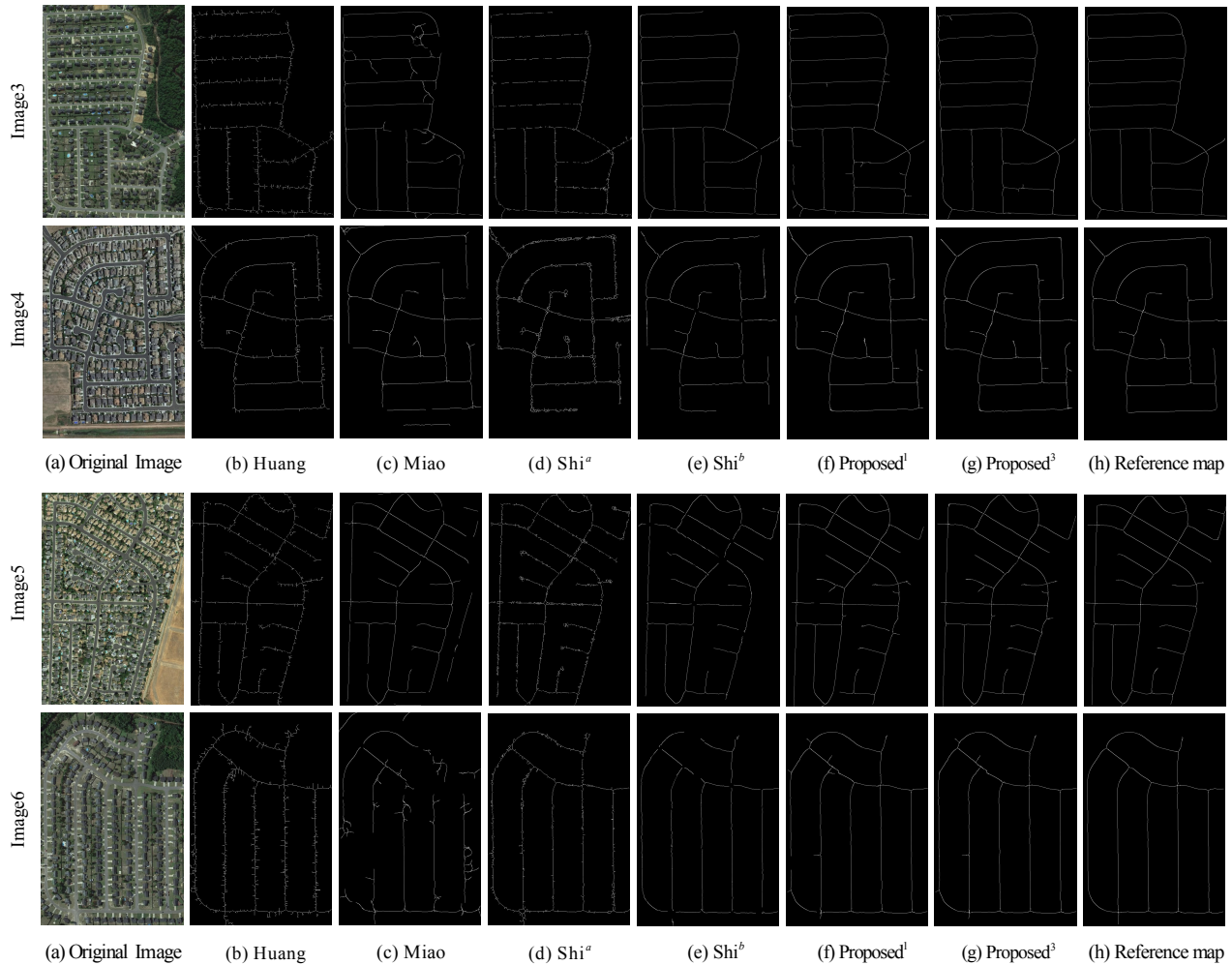}}
\caption{Visual comparisons of centerline extraction results on Data $\#1$. There are four rows and eight columns of subfigures. Each row shows the results of one image. From left to right: (a) original image, (b) result of Huang \cite{journals_ijrs_Huangxin09}, (c) result of Miao \cite{journals_lgrs_MiaoSZW13}, (d) result of $\textrm{Shi}^{a}$ \cite{journals_lgrs_ShiMWZ14}, (e) result of $\textrm{Shi}^{b}$ \cite{journals_tgrs_ShiMD14}, (f) result of $\textrm{Proposed}^{1}$, (g) result of $\textrm{Proposed}^{3}$,  (b) the reference map.}
\label{centerline_compare1}
\end{figure*}
\begin{figure*}[t]
\centerline{\includegraphics[width=17cm]{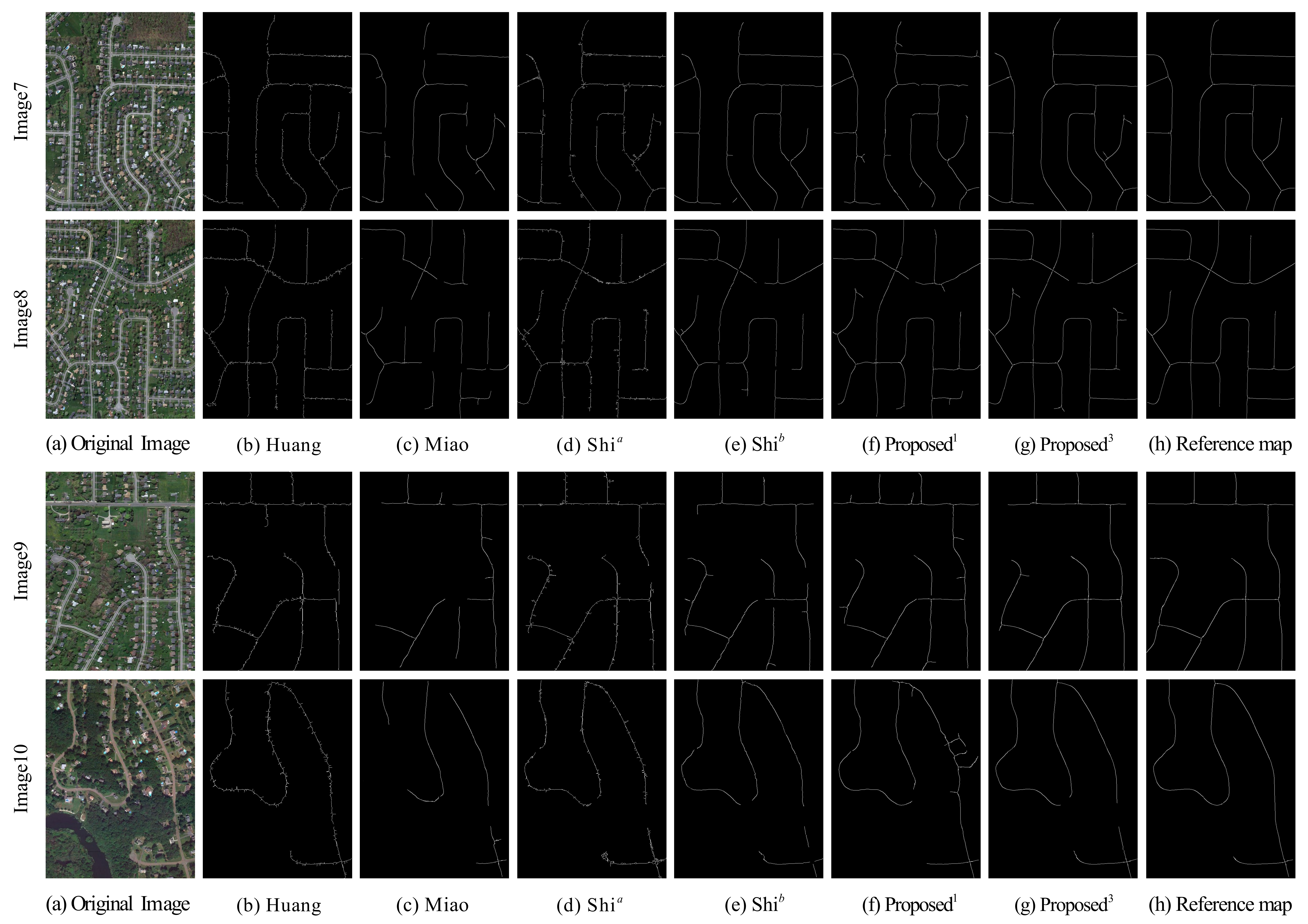}}
\caption{Visual comparisons of centerline extraction results on Data $\#2$. There are eight rows and four columns of subfigures. Each row shows the results of one image. From left to right: (a) original image, (b) result of Huang \cite{journals_ijrs_Huangxin09}, (c) result of Miao \cite{journals_lgrs_MiaoSZW13}, (d) result of $\textrm{Shi}^{a}$ \cite{journals_lgrs_ShiMWZ14}, (e) result of $\textrm{Shi}^{b}$ \cite{journals_tgrs_ShiMD14}, (f) result of $\textrm{Proposed}^{1}$, (g) result of $\textrm{Proposed}^{3}$,  (b) the reference map.}
\label{centerline_compare2}
\end{figure*}
\subsection{\bf Parameter setting}
There are mainly two parameters in the proposed method, i.e. the balancing factor $\alpha$ in GC and the scale facor $\sigma$ in TV. In the experiments, we set other parameters as follows: The three different number of superpixels are set as $8000$, $10000$ and $12000$ in $\textrm{Proposed}^{3}$, while in $\textrm{Proposed}^{1}$ we choose the number of superpixels as 10000 in the case of no other special instructions. In MJCR, we randomly choose $50$ positive samples and negative samples from the ground truth. To obtain a sparse coefficients $\bm{\alpha}^*$, we set the weighting factor $\lambda=10$. We set the window size of neighborhood as $2\sigma$ in NMS. In the fitting based connection method, the width of searching window R is set to be $\sigma$.  For Data $\#$1, the road width is about $12$-$15$ pixels, thus we choose 3 as the buffer width. While for Data $\#2$ we set the buffer width $\rho$ as $2$ pixels. Actually, in the experiments, we find that satisfactory results can be obtained for most images when setting $\alpha=0.8$ in GC. Thus we keep this parameter fixed in the following experiments. For the scale factor $\sigma$ in TV, we get the optimal value for each image via cross validation, which will be discussed later.
\subsection{\bf Comparison of road area extraction results}
In order to illustrate an intuitive comparison of different methods on road area extraction, we display the results by comparing the extracted map with the reference map in Fig.~\ref{classification_compare}. $\textrm{Shi}^{a}$ and $\textrm{Shi}^{b}$ achieve better performance than other two comparing methods, while both of them are inefficient to extract the road under the occlusions of cars and trees. Huang's method can extract the complete road network well. However, it brings in more false positives (the red areas in Fig. \ref{classification_compare}(\textcolor{blue}{b})). For Image2, Miao's method get unsatisfactory result. This is because the homogeneous regions are considered to be road after the edge-filtering method \cite{journal_isprs_Bakker02}, meanwhile in Image2 some forest areas are also homogeneous regions too. Thus it is hard to select an appropriate threshold to distinguish them from the road areas. From the comparing results, it shows that the proposed method extracted more satisfactory and coherence road map, especially for Image2. Besides, our approach is more robust against the occlusions of trees and cars.
\par The quantitative performances of road area extraction results are summarized in Tables~\ref{table_classification}. As we know, COM and COR can be one-sided metrics. We can enlarge one at the cost of the other one. The Q term combines both of them, thus is a overall measurement. As can be seen from the table, our method achieve the highest value in Q. $\textrm{Shi}^{a}$ and $\textrm{Shi}^{b}$ obtain higher values than Huang's method and Miao's method by a large margin. In summary, the Q value of our method is $4\%$ higher than second best method, which demonstrates the validity of our approach on the road area extraction.
\subsection{\bf Comparison of road centerline result on Data $\#1$}
\par We carried out experiments on all the $30$ images in Data $\#1$. Due to space limit, we select $4$ of them to illustrate the comparing results among the aforementioned methods in Fig.~\ref{centerline_compare1}. By comparing the extracted road centerline map with its corresponding reference map, we can have following conclusions: (1) Through morphological thinning based road centerline method (e.g. Huang's method and $\textrm{Shi}^{a}$'s method) is fast and easy to extract road centerlines, this method produces small spurs, which reduce the smoothness and correctness of road network. (2) The regression based road centerline methods (e.g. Miao's method and $\textrm{Shi}^{b}$'s method) can extract relatively smooth centerlines well. However, these methods can't link the centerlines well in the intersection areas (see subfigure (\textcolor{blue}{c}) and (\textcolor{blue}{e}) of Image5 in Fig.~\ref{centerline_compare1}). (3) There are some false positives in Miao's method, this is because some forest areas and bare soil are also tend to be homogenous regions in VHR remote sensing images, thus it is hard to choose an appropriate threshold to distinguish them with road areas. (4) Our proposed methods ($\textrm{Proposed}^{1}$ and $\textrm{Proposed}^{3}$) achieve more smooth and continuous road centerlines than other comparing methods, and both of them work well in the junction areas. Besides, the centerline result of $\textrm{Proposed}^{3}$ is more similar to the reference map than $\textrm{Proposed}^{1}$, which demonstrates that multi-scale information fusion is suitable for the road centerline extraction task.
\par Quantitative comparisons of road centerline extraction result among different methods are illustrated in Table \ref{centerline_table1}. As we shall see, in terms of COR, the proposed method ($\textrm{Proposed}^{3}$) achieves the second best results after $\textrm{Shi}^{b}$'s method, while in terms of COM, the proposed method obtains bigger values than $\textrm{Shi}^{b}$'s method by a large margin. Thus our proposed method achieves relatively higher Q values, which is an overall evaluation index. The thinning based methods (i.e. Huang's method and $\textrm{Shi}^{a}$'s method) and Miao's method give relatively low values of COR and Q due to the appearance of false positives and small spurs. $\textrm{Proposed}^{1}$ obtains the second best performance among all the methods, while it is inferior to the $\textrm{Proposed}^{3}$ almost in each comparing items.
\subsection{\bf Comparison of road centerline result on Data $\#2$}
\par Visual comparisons of centerline extraction results on Data $\#2$ among different methods are displayed in Fig.~\ref{centerline_compare2}. we find that there are large areas of dense forests and grasslands in those images, which is hard to extract correct road network from them for Miao's method. Thus the centerline result of Miao's method is incomplete (see Fig.~\ref{centerline_compare2}(\textcolor{blue}{c})). Huang's method and $\textrm{Shi}^{a}$'s method can extract relatively complete road network, while their methods produce small spurs due to the algorithm of centerline extraction. $\textrm{Shi}^{b}$'s method and $\textrm{Proposed}^{1}$  achieve better performance than other methods (e.g. Huang's method, Miao's method and $\textrm{Shi}^{a}$'s method), while both of them are inferior to the $\textrm{Proposed}^{3}$ in term of completeness and smoothness.
\par Table~\ref{centerline_table2} presents the quantitative comparisons on Data $\#2$ for different methods. As we shall see, Miao's method achieves low values both in terms of COM and Q, because the road network is hard to extract under the conditions of large areas of forests and grasslands. Huang's method and $\textrm{Shi}^{a}$'s method gain relatively large values in COM and small values in COR, which leads to small Q values. $\textrm{Shi}^{b}$'s method and $\textrm{Proposed}^{1}$ achieve similar performance in average among all the four images, while $\textrm{Proposed}^{3}$ achieves higher values than both of them in terms of all the three metrics (e.g. COM, COR and Q). Specifically, the average Q value among all the images of the $\textrm{Proposed}^{3}$ is $3\%$ higher than the second best method.
\subsection{\bf Impact of parameter $\sigma$}
\begin{figure}[t]
\centerline{\includegraphics[width=7.5cm]{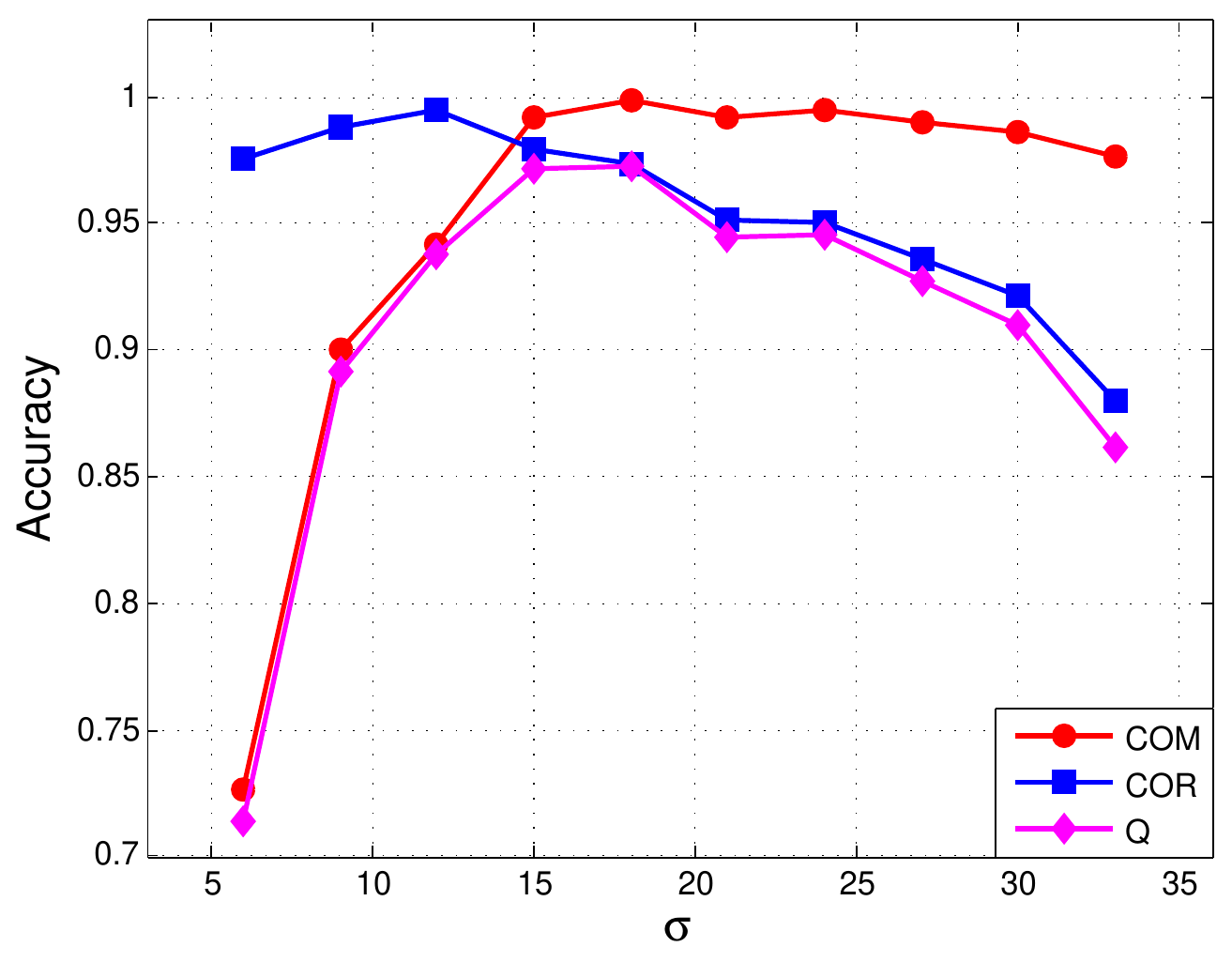}}
\caption{Quantitative comparisons of centerline extraction results with different scale parameters. COM, COR and Q are calculated when the buffer with $\rho=2$. (Best viewed in color)}
\label{sigma_line_plot}
\end{figure}
\begin{table*}[t]
\caption{Quantitative comparisons with state-of-the-art methods in Data $\#2$, where the {\color{red}{\textbf{red values}}} marked in bold are the best and the bold {\color{blue}{\textbf{blue values}}} are the second best. (Best viewed in color)
\label{centerline_table2}}
\centerline{\includegraphics[width=17cm]{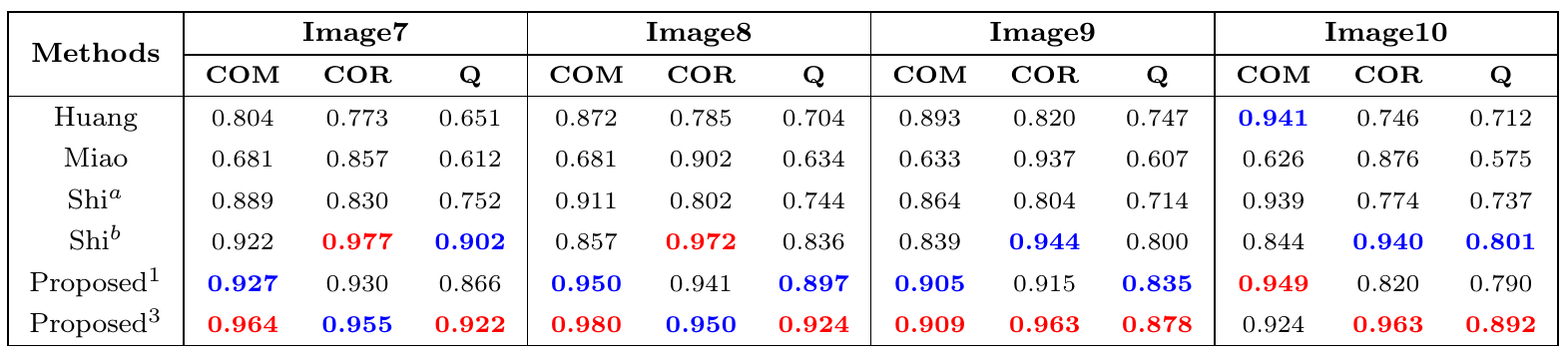}}
\end{table*}
\begin{figure*}[t]
\centerline{\includegraphics[width=17cm]{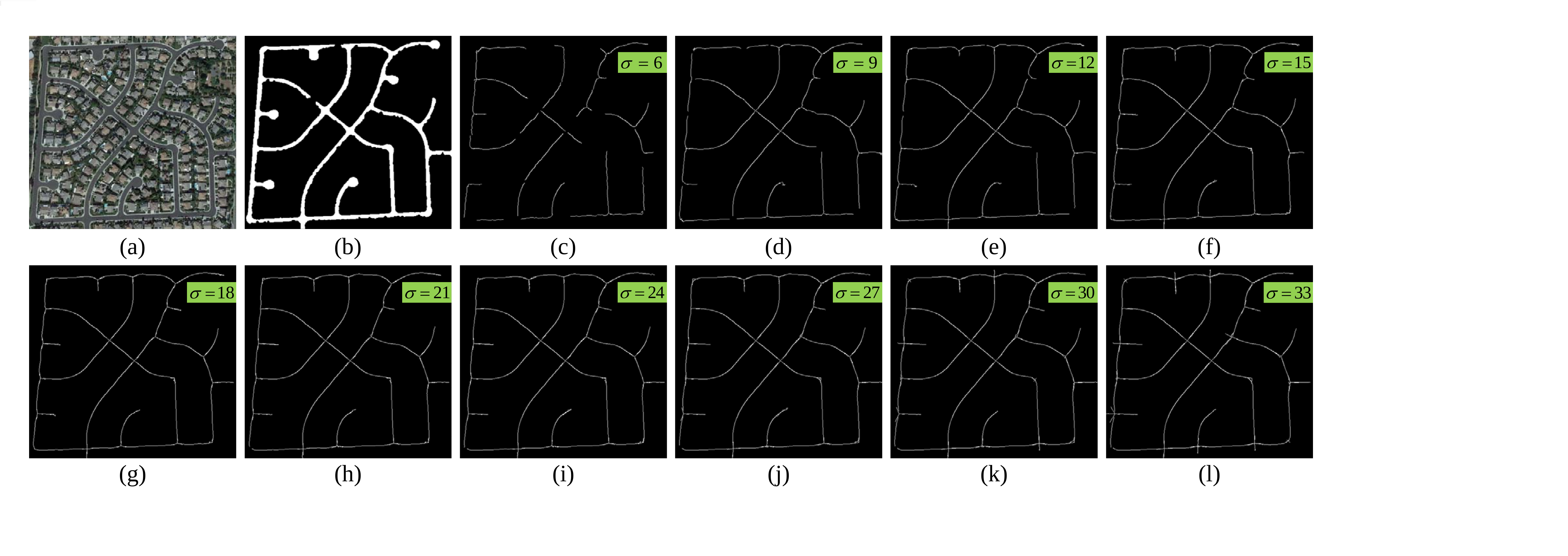}}
\caption{Visual comparisons with different scale parameters for the proposed method. There are two rows and six columns of subfigures. From left to right and up to down: (a) original image, (b) classification result of the $\textrm{Proposed}^{3}$, (c) centerline result with $\sigma=6$, (d) centerline result with $\sigma=9$, (e) centerline result with $\sigma=12$, (f) centerline result with $\sigma=15$, (g) centerline result with $\sigma=18$, (h) centerline result with $\sigma=21$, (i) centerline result with $\sigma=24$, (j) centerline result with $\sigma=27$, (k) centerline result with $\sigma=30$, (l) centerline result with $\sigma=33$.}
\label{sigma_adjust}
\end{figure*}

\par In the proposed method, tensor voting is used to extract road centerlines. There is only one main parameter $\sigma$ to be tuned in TV. In addition, the window size in NMS and the width of searching window in fitting based connection algorithm are also determined by $\sigma$. Thus it is important to choose an appropriate $\sigma$ for each image. To give an instructions for users to choose a suitable $\sigma$, the sensitivity of $\sigma$ is tested in this section.
\par Visual comparisons with different $\sigma$ values are illustrated in Fig.~\ref{sigma_adjust}. As is seen from the figure, there are some discontinuous regions in the classification result (see Fig.~\ref{sigma_adjust}(\textcolor{blue}{b})) due to the occlusions of trees (see Fig.~\ref{sigma_adjust}(\textcolor{blue}{a})). One reason to choose TV as centerline extraction algorithm is that TV can eliminate the road discontinuity with a suitable scale parameter. When $\sigma$ is small, TV is not only ineffective to link the discontinuous areas, but also can't extract centerlines well for those continuous regions (see Figs.~\ref{sigma_adjust}(\textcolor{blue}{c}), \ref{sigma_adjust}(\textcolor{blue}{d}) and \ref{sigma_adjust}(\textcolor{blue}{e})). With the increase of scale parameter $\sigma$ within a certain range, TV can extract centerlines for the connected areas and link the centerlines for the disconnected areas (see Figs.~\ref{sigma_adjust}(\textcolor{blue}{f}), \ref{sigma_adjust}(\textcolor{blue}{g}), \ref{sigma_adjust}(\textcolor{blue}{h}), \ref{sigma_adjust}(\textcolor{blue}{i}) and \ref{sigma_adjust}(\textcolor{blue}{j})). When $\sigma$ is greater than a certain value, although TV can connect the discontinuous centerlines well, it also brings in some false positives (see Figs.~\ref{sigma_adjust}(\textcolor{blue}{k}) and \ref{sigma_adjust}(\textcolor{blue}{l})). Those false positives lead to wrong navigation information in practical application. Therefore, to get a smooth and complete road network, a suitable value for scale parameter should be carefully selected.
\par To give an intuitive comparison, the quantitative performances with different scale parameters are summarized in Fig.~\ref{sigma_line_plot}. In this figure, COM, COR and Q are calculated with the buffer width $\rho=2$. When $\sigma$ is less than $12$ pixels, COM, COR and Q are all increase along with the growth of $\sigma$. Then with the increase of $\sigma$, COR declines gradually, the overall metric (Q) reduces even though COM remains stable. We find that satisfactory results can be obtained with $\sigma$ in a large range (e.g. $15$-$24$ pixels). Specifically, the highest Q values are obtained when $\sigma$ is set to be $15$ or $18$ pixels. It should be noted that the road width $w$ of the test image is about $12$ pixels, thus it is suitable to set the scale parameter as $\sigma=1.5w$.
\section{Conclusions}
\label{Section5}
\par In this paper, we propose an accurate road centerline extraction method for VHR remote sensing images. The proposed method contains three stages: homogeneous road area segmentation, smooth and correct road centerline extraction, and centerline connection around the road intersections. Specifically, in the $1^{\textrm{st}}$ stage, to obtain an accurate road area extraction result, multiscale joint collaborative representation and graph cuts are introduced to incorporating the mutiscale features and spatial information. In the $2^{\textrm{nd}}$ stage, to overcome the shortcomings of morphological thinning algorithm, tensor voting and non-maximum suppression algorithm are utilized to extract smooth and correct road centerlines. Finally, in the $3^{\textrm{rd}}$ stage, to tackle the ineffectiveness of the existing methods around the road intersections, a fitting based centerline connection algorithm is proposed to complete the road network. In terms of both quantitative and visual performances, the proposed method achieves better results than all the other comparing methods. Moreover, the proposed method are not sensitive to the parameters, which are tuned easily.  As another contribution, a new and challenging road centerline extraction dataset for VHR remote sensing images will be publicly available for further studies.


%





\ifCLASSOPTIONcaptionsoff
  \newpage
\fi



\balance
\bibliographystyle{IEEEtran}
\bibliography{ref}
\end{document}